\documentclass{article}

\PassOptionsToPackage{numbers}{natbib}
 \usepackage[preprint]{tsrml_2022}

\usepackage[utf8]{inputenc} % allow utf-8 input
\usepackage[T1]{fontenc}    % use 8-bit T1 fonts
\usepackage[colorlinks]{hyperref}       % hyperlinks
\usepackage{url}            % simple URL typesetting
\usepackage{booktabs}       % professional-quality tables
\usepackage{amsfonts}       % blackboard math symbols
\usepackage{nicefrac}       % compact symbols for 1/2, etc.
\usepackage{microtype}      % microtypography
\usepackage{xcolor}         % colors
\usepackage[thinlines]{easytable}
\usepackage{subcaption}
\usepackage{multirow} 
\usepackage{todonotes}
\usepackage{epsfig}
\usepackage{times}
\usepackage{graphicx}
\usepackage{amsmath}
\usepackage{amssymb}
\usepackage{amsthm}
\usepackage{latexsym}

\title{Evaluating Trade-offs in Computer Vision Between Attribute Privacy, Fairness and Utility}

\author{%
  William Paul \\
  JHU/APL\\
  \And
  Philip Mathew\\
  JHU/APL\\
  \And
  Fady Alajaji\\
  Queens University\\
  \And 
  Philippe Burlina\\
  JHU/APL\\
}

\newtheorem{definition}{Definition}

\begin{document}

\maketitle

\begin{abstract}
This paper investigates to what degree and magnitude tradeoffs exist between utility, fairness and attribute privacy in computer vision. Regarding privacy, we look at this important problem specifically in the context of attribute inference attacks, a less addressed form of privacy. To create a variety of models with different preferences, we use adversarial methods to intervene on attributes relating to fairness and privacy. We see that that certain tradeoffs exist between fairness and utility, privacy and utility, and between privacy and fairness. The results also show that those tradeoffs and interactions are more complex and nonlinear between the three goals than intuition would suggest.
\end{abstract}

\section{Introduction}

Despite recent successes in developing artificial intelligence (AI) and deep learning (DL) systems~\cite{lecun2015deep} for a range of applications including image interpretation, vehicular navigation, robotics, and medicine~\cite{he2016deep,Ren2015FasterRT,Redmon2015YouOL,ronneberger2015u,burlina2019automated,mnih2015human,silver2017mastering}, ensuring these systems are reliable enough to build trust is still an open problem. Trust and ethical concerns about AI systems such as explainability,  adversarial vulnerabilities (adversarial machine learning or AML)~\cite{goodfellow2014explaining,carlini2017adversarial} and importantly, fairness~\cite{burlina2020addressing,burlina2020low} and privacy~\cite{shokri2017membership,paul2021defending,hui2021practical} -- the two main foci of this work --  have recently put into question the deployment of certain autonomous systems. 

A common theme underpinning a number of these concerns is being able to discern what information the model is using to make its decision. Fairness across groups, which is trying to ensure performance of the system is equal in some respect between different subpopulations denoted by some sensitive attribute, is typically viewed as the model learning correlations between the sensitive attribute and the task in an excessive manner. Privacy has a number of different forms, but commonly focuses on cases either where the model may memorize the training data used in some manner or, what we address in this work, where features are used as a surrogate for the original data to infer private attributes about individuals, known as attribute privacy. For attribute privacy, the model effectively carries information about the private attribute into the features, either inadvertently or for use at the task at hand, acting similarly to concerns raised in addressing fairness. However, if we want to address these concerns simultaneously, there can be a balancing act for determining what information is used for the task, leading to potential trade-offs between how well the task is performed (utility) as well as the fairness and attribute privacy of the model. Though such trade-offs have been shown for differential privacy \cite{jagielski2019differentially}, to the best of our knowledge, we have not seen such an analysis of trade-offs between all three areas for attribute privacy or computer vision related tasks such as facial recognition.

Consequently, in this work, we seek to explore and evaluate how these trade-off present themselves, how severe they are, as well as to what degree we can control them. We leverage adversarial methods as one of the most prominent methods to address group fairness, also using these methods for attribute privacy due to similar motivations. We evaluate on a variety of image datasets, focusing on the tasks of classification and facial recognition, and train a diverse family of models with different preferences of each area of concern to better understand their corresponding metrics are affected.

\begin{figure*}
    \centering
    \includegraphics[width=\linewidth]{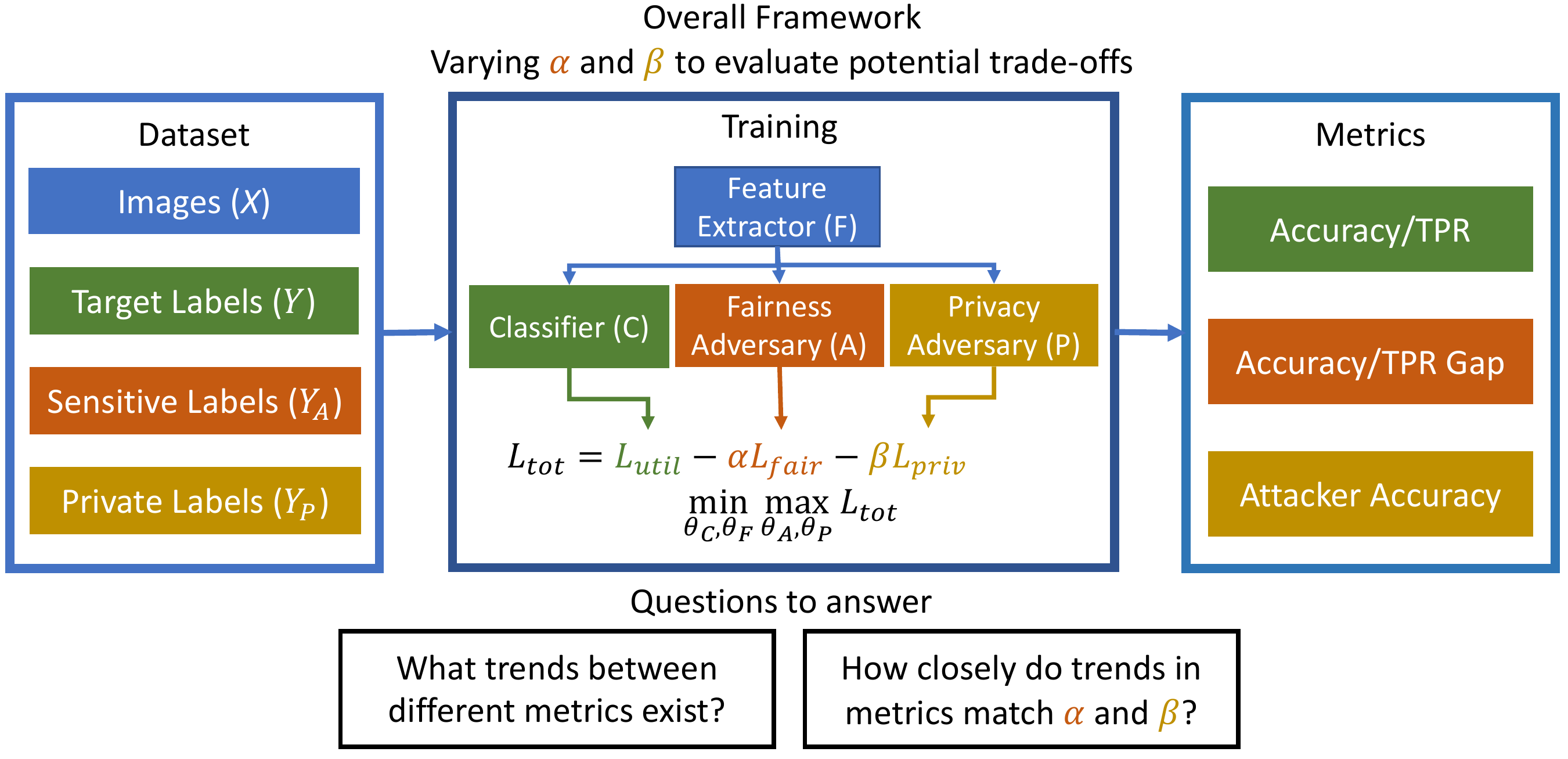}
    \caption{Diagram depicting our overall methodology and framework. In order to create classifiers with different characteristics, we modulate the amount of information related to the sensitive and private attributes. Finally, we evaluate utility, fairness, and privacy holistically, answering what interactions exist between them as well as how they interact with the chosen coefficients.}
    \label{fig:overview}
\end{figure*}

\section{Prior Work}

There has been increased interest in looking at aspects of assured AI now that deep learning has demonstrated it can operate on par or beyond human capabilities for tasks including classification, detection in natural images and that machine learning model allow for having similar performance to that of clinicians for diagnostics for pathologies like skin lesions or retinopathies \cite{burlina2019assessment}. There are a number of different areas within assuring AI, such as being able to provide explanations, robust to imperceptual attacks \cite{carlini2017adversarial}, as well as our focus, ensuring fair and private operation.

Fairness is typically viewed disparities between subpopulations characterized by some attribute, otherwise known as group fairness, and has been investigated in a number of prior studies ranging from natural language processing to medical imagery \cite{zemel2013learning,bolukbasi2016man,zemel2013learning,kinyanjui2020fairness}. A broad taxonomy of methods~\cite{caton2020fairness} is organized along the line of approaches that intervene in different parts of the training process, from data preprocessing, calibrating the predictions, and changing the model, including modifying the loss function used which is our approach in this work. Model interventions are a common way to address fairness \cite{goodfellow2014generative,beutel2017data,alvi2018turning}, typically focusing on reducing the undue influence of an attribute within some internal features of the model. This is done via reducing some notion of distance between features of different subpopulatons, such as maximum mean discrepancy, or other form of discrimination such as those induced by adversaries \cite{wadsworth2018achieving,zhang2018mitigating,song2019learning}. 

Privacy has a number of different instantiations, such as protecting against membership inference attacks, where an attacker is trying to infer whether a data point was used for training, or achieving some level of differential privacy, controlling how much influence the training dataset has on the final weights. However, our focus in this work is on a third form called attribute inference attacks, where an attacker is attempting to predict a private attribute about the original data from a transformed or masked version acting as a surrogate. In information theory, attempting to ensure robustness to this attack while maintaining some level of utility of the transformed data is known as the privacy funnel \cite{makhdoumi2014information,asoodeh2014funnel}. More specifically, the goal of the privacy funnel is to maximize the mutual information between the transformed data and the original data, while keeping the mutual information between transformed data and the private attribute below a certain threshold. Unlike differential privacy, attribute privacy requires the network to actively prune out information about the private attribute in order to defend against an adversarial threat model. Consequently, there is significant overlap between addressing attribute privacy and group fairness, as both arguably only differ in terms of how the included information in the feature is presented downstream \cite{barbano2021}.

Addressing both group fairness and attribute privacy simultaneously is less studied however. For differential privacy, \cite{jagielski2019differentially} appears to be the most prominent approach towards addressing both, though this work focuses on differential privacy.  \cite{robinson2021fairnessprivacyfacial, dhar2021pass} both evaluated how well adversarial learning in facial recognition can help ensure group fairness and attribute privacy. Motivating our work however, they did not investigate what trends exist between fairness, privacy, and utility.
\section{Methodology}
We detail notation and defintions used, how we construct our framework for experiments, as well as what metrics are used. 

\textbf{Notation and Definitions:} In terms of notation used, we consider the dataset to be comprised of images $X$, target labels $Y$,predicted labels $\hat{Y}$, sensitive labels $Y_A$, and private labels $Y_P$. The models we use consist of: the feature extractor F, converting images to features with weights $\theta_F$; the classifier (C) using the features to predict $Y$ with weights $\theta_C$; the fairness adversary A attempting to predict $Y_A$ from the features using weights $\theta_A$; and the privacy adversary (P)  attempting to predict $Y_P$ from the features using weights $\theta_P$. $\alpha$ and $\beta$ are linear coefficients scaling the loss for the fairness and privacy adversaries. 

We review the fairness criteria we focus on below:

\begin{definition}[Fairness Criteria]
    Given the sensitive label $A$, target label $Y$, and the classifier predictions $\hat{Y}$, the classifier is said to satisfy:
    \begin{itemize}
        \item accuracy parity if $P(\hat{Y} = Y | Y_A) = P(\hat{Y}=Y)$, i.e., the predictions matching Y are independent of A.
        \item equality of opportunity if for a given Y=y, $P(\hat{Y} | Y_A, Y=y) = P(\hat{Y} | Y=y)$, i.e., the predictions for the class is y, commonly taken to be the positive class, are independent of A.
    \end{itemize}
\end{definition}

Demographic parity and equality of odds are two notable criteria we do not evaluate on. For demographic parity, all of the tasks we consider in this work focus on more descriptive attributes which are less likely to have allocative bias that demographic parity addressed, where the actual labeling assignment is not fair. Similarly, accuracy parity is focused on as a weaker form of equality of odds, not penalizing the model trading false negatives for false positives.

For attribute privacy, we consider the following threat model for attribute privacy:
\begin{definition}(Attribute Privacy) 
Given oracle access to $F(X)$ for input data $X$, producing features associated with $X$, access to labeled public data $\{(X^i, Y_P^i)\}_{i=1,\dots,n}$, and access to $\{F(\tilde{X}^i)\}_{i=1,\dots,\tilde{n}}$, the features corresponding to hidden data $\tilde{X}$, an adversary is trying to infer on the hidden data the unknown $\tilde{Y}_P$ corresponding to $\tilde{X}$ better than chance, i.e., achieving $P(\tilde{Y}_P | F(X)) > P(\tilde{Y}_P)$.
\end{definition}
The adversary's objective is to generalize from the features they have corresponding private attributes for to the features. This form of privacy is not addressed by usual forms of differential privacy, which try to lessen the influence of the training data used, as the feature extractor must activity prune information about $Y_P$ from $X$.

\subsection{Investigating Trade-Offs between Fairness and Privacy} 
There are a number of potential methods that can remove information about an attribute from features, but a consistently used technique is using adversaries to remove information. Consequently, when training models, the corresponding optimization is:
\begin{equation}
    \min_{\theta_C,\theta_F} \max_{\theta_A, \theta_P} \mathbb{CE}(Y, C(F(X))) - \alpha \mathbb{CE}(Y_A, A(F(X), Y))  - \beta \mathbb{CE}(Y_P, P(F(X), Y)) 
\end{equation}

where $\mathbb{CE}$ denotes the cross entropy and both adversaries are conditional on $Y$ as we are targeting fairness criteria that are conditional on $Y$ and for privacy as the attacker can easily acquire $Y$ from the features.

For each dataset, as shown in Figure \ref{fig:overview}, we then perform a grid search over $\alpha$ and $\beta$, with each hyperparameter either zero or going from $10^{-2}$ to $10$ logarithmically in 10 steps, and evaluate the corresponding effects on utility, fairness and privacy.

\begin{table}[t]
\centering
\scriptsize
    \centering
    
    \captionof{table}{Results on single metrics on different datasets. We show the baseline metrics without any intervention, where the specific metric for utility and fairness is shown in parentheses for those columns. For privacy, the percentage in parentheses is when the attack accuracy is no better than chance.}
    	\begin{tabular}{ccccccc}
	    \toprule
        Dataset& Baseline Utility& Baseline Fairness& Baseline Privacy& Best Utility& Best Fairness& Best Privacy \cr
        \midrule
        \midrule
        CelebA &74.26\% (Acc.)&9.27\% (Acc. Gap)&82.54\% (50\%)&76.26\% & 4.11\% & 70.33\%\cr
        CelebA FR &85.69\% (TPR)&7.53\% (TPR Gap)&69.92\% (50\%) & 85.69\% & 1.87\% & 65.38\% \cr 
        EyePACs&63.33\% (TPR)&21.00\% (TPR Gap)&79.83\% (33\%) & 70.00\% & 8.00\% & 67.77\% \cr
        CheXpert&62.69\% (TPR)&24.95\% (TPR Gap)&38.69\% (25\%) & 67.16\% & 14.57\% & 33.47\%  \cr 
        
        \bottomrule
	\end{tabular}
	
    \label{tab:single}
\end{table}
\begin{table*}[t]
\centering
\scriptsize
    \centering
    
    \captionof{table}{Metrics incorporating multiple metrics. We negate utility when computing correlations on utility to match direction of improvement. For CSR metrics, letters in parantheses denote which grouping of $(\alpha,\beta)$ the ranking belongs to.}
    	\begin{tabular}{ccccccc}
	    \toprule
        Dataset& U./F. Corr.& U./P. Corr.& F./P. Corr. & $\text{CSR}(0.6, 0.2, 0.2) $  & $\text{CSR}(0.2, 0.6, 0.2) $   & $\text{CSR}(0.2, 0.2, 0.6) $   \cr
        \midrule
        \midrule
        CelebA & 0.50&-0.01&-0.14 & 91.04\% (H., H.) & 88.98\% (H., H.) & 88.39\% (H., H.)\cr
        CelebA FR &-0.92&-0.08& 0.15 & 77.67\% (L., M.) & 75.11\% (H., M.) & 77.64\% (M., M.) \cr 
        EyePACs&0.58&-0.25&-0.19 & 76.00\% (H., M.) & 92.00\% (H., M.) & 92.00\% (H., M.)\cr
        CheXpert&-0.02&0.07&0.03 & 87.71\% (M., M.) & 82.42\% (H., L.) & 87.71\% (M., M.)  \cr 
        \bottomrule
	\end{tabular}
	
    \label{tab:tradeoffs}
\end{table*}
\textbf{Metrics:} We utilize several metrics in this study, primarily for evaluating area of concern as well as evaluating a combination of these metrics. For utility, we use either the overall accuracy or true positive rate. For fairness, to expand into cases where $Y_A$ is not binary, we take fairness to be the maximum pairwise absolute difference of the utility metric chosen over the subpopulations defined by $Y_A$. For privacy, to match the threat model, we train an separate adversary, a linear model in this work, on features extracted from the validation dataset, mimicking the access to the labeled public data. The linear model is trained using loss re-weighting to ensure the adversary does not be a constant prediction. The hidden data is consequently the test dataset used, and the metric for privacy is the balanced accuracy for reasons noted in the previous sentence.

For metrics evaluating the combination of metrics, we use pairwise correlation to measure the linear relationship between different metrics. To incorporate preferences that are not strictly pairwise, we also introduce a metric called Conjunctive Soft Ranking (CSR) to rank the models. CSR is effectively the convex combination between the normalized metrics for each area, normalized so that the worst model over $\alpha$ and $\beta$ is $0\%$ and the best is $100\%$. 
\begin{gather}
    N(M_{\alpha, \beta}) = \frac{M_{\alpha, \beta} - \min_{\alpha', \beta'}M_{\alpha', \beta'}}{\max_{\alpha', \beta'}M_{\alpha', \beta'} - \min_{\alpha', \beta'}M_{\alpha', \beta'}} \\
    CSR_{\alpha, \beta}(\gamma_U, \gamma_A, \gamma_P) = 100(\gamma_U N(M_{\alpha, \beta}^U) + \gamma_A (1 - N(M_{\alpha, \beta}^A)) + \gamma_P (1 - N(M_{\alpha, \beta}^P)))
\end{gather}

where $\gamma_U$,$\gamma_A$,and $\gamma_P$ sum to one, and $M_{\alpha,\beta}$ denotes the metric for the model trained using $\alpha$ and $\beta$, with the superscript denoting which area the metric is measuring. To probe these rankings and see how the best $\alpha$ and $\beta$ changes as the preference changes, we chose preferences focusing on a primary metric (using a weight of 0.6) and weighing the other two equally (using a weight of 0.2 each).

\section{Experiments}
\begin{figure*}[!t]
\centering
\begin{subfigure}{.33\textwidth}
  \centering
  \includegraphics[width=\linewidth]{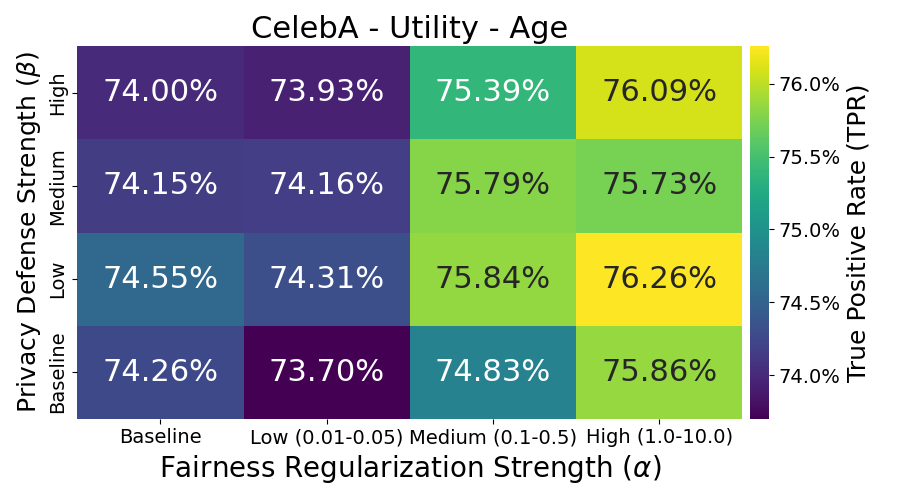}
  \label{fig:sub1}
\end{subfigure}%
\begin{subfigure}{.33\textwidth}
  \centering
  \includegraphics[width=\linewidth]{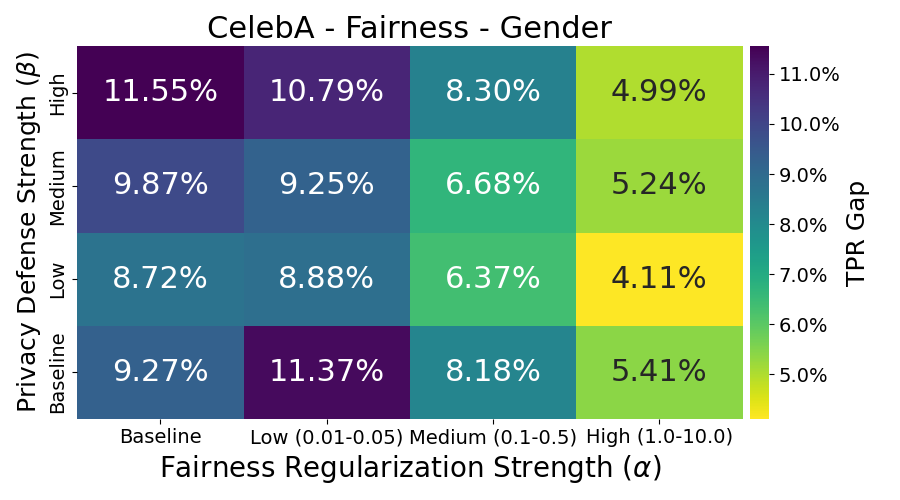}
  \label{fig:sub2}
\end{subfigure}
\begin{subfigure}{.33\textwidth}
  \centering
  \includegraphics[width=\linewidth]{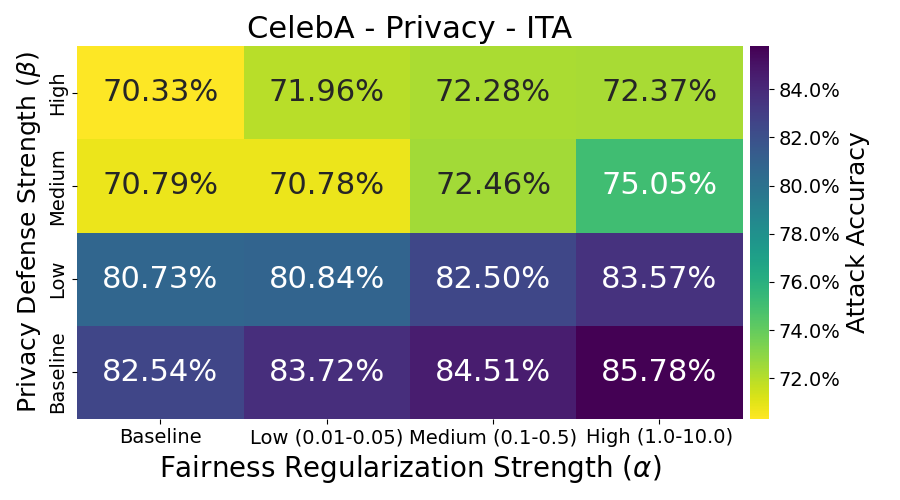}
  \label{fig:sub2}
\end{subfigure}
% \begin{subfigure}{.32\textwidth}
%   \centering
%   \includegraphics[width=\linewidth]{figures/celeba_test_heatmap_cai_range_0.2_0.2.png}
%   \label{fig:sub1}
% \end{subfigure}%
% \begin{subfigure}{.32\textwidth}
%   \centering
%   \includegraphics[width=\linewidth]{figures/celeba_test_heatmap_cai_range_0.6_0.2.png}
%   \label{fig:sub2}
% \end{subfigure}
% \begin{subfigure}{.32\textwidth}
%   \centering
%   \includegraphics[width=\linewidth]{figures/celeba_test_heatmap_cai_range_0.2_0.6.png}
%   \label{fig:sub2}
% \end{subfigure}
\vspace{-0.5cm}
\caption{Heatmaps on CelebA over the different grouped regularization strengths for $\alpha$ and $\beta$.}
\label{fig:celeba}
\end{figure*}
\begin{figure*}[!t]
\centering
\begin{subfigure}{.33\textwidth}
  \centering
  \includegraphics[width=\linewidth]{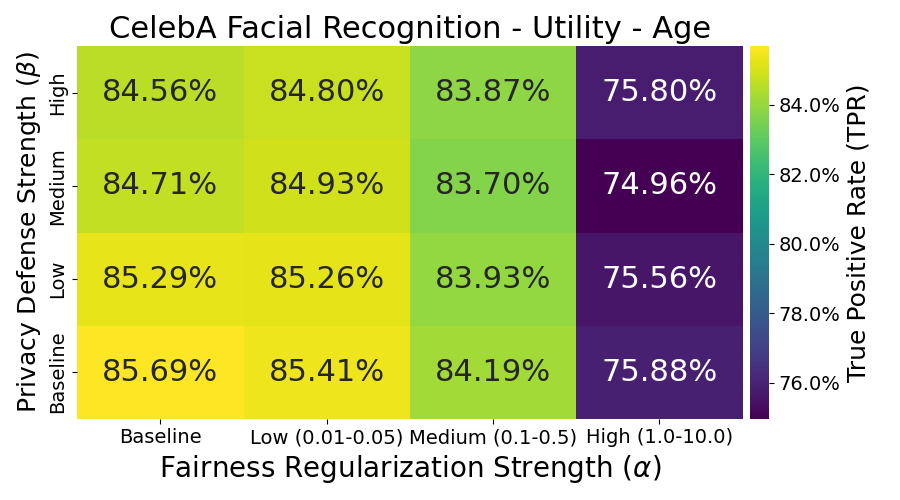}
  \label{fig:sub1}
\end{subfigure}%
\begin{subfigure}{.33\textwidth}
  \centering
  \includegraphics[width=\linewidth]{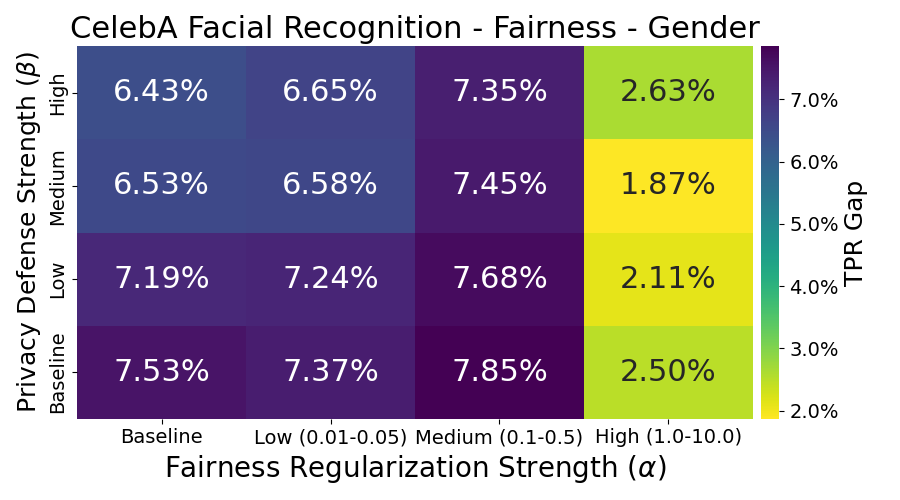}
  \label{fig:sub2}
\end{subfigure}
\begin{subfigure}{.33\textwidth}
  \centering
  \includegraphics[width=\linewidth]{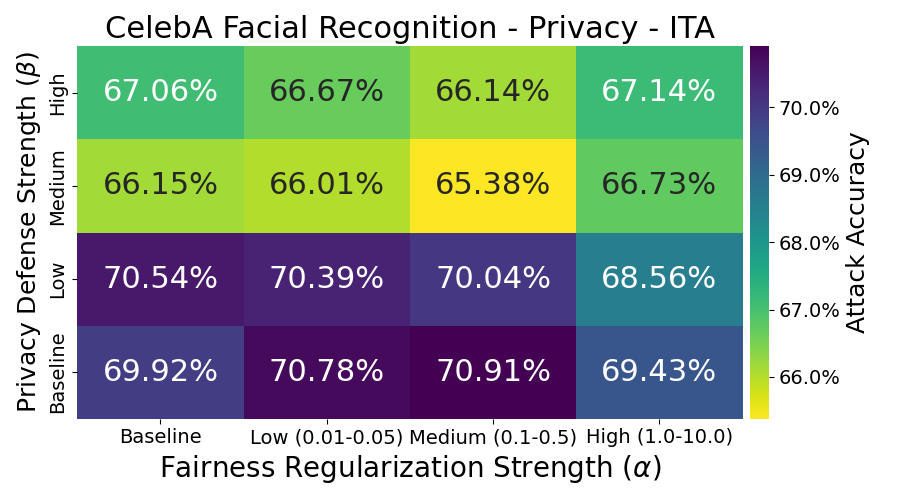}
  \label{fig:sub2}
\end{subfigure}
% \begin{subfigure}{.32\textwidth}
%   \centering
%   \includegraphics[width=\linewidth]{figures/celeba_fr_test_heatmap_cai_range_0.2_0.2.png}
%   \label{fig:sub1}
% \end{subfigure}%
% \begin{subfigure}{.32\textwidth}
%   \centering
%   \includegraphics[width=\linewidth]{figures/celeba_fr_test_heatmap_cai_range_0.6_0.2.png}
%   \label{fig:sub2}
% \end{subfigure}
% \begin{subfigure}{.32\textwidth}
%   \centering
%   \includegraphics[width=\linewidth]{figures/celeba_fr_test_heatmap_cai_range_0.2_0.6.png}
%   \label{fig:sub2}
% \end{subfigure}
\vspace{-0.5cm}
\caption{Heatmaps on CelebA for facial recognition over the different grouped regularization strengths for $\alpha$ and $\beta$.}
\label{fig:celebafr}
\end{figure*}
We evaluate on CelebA \cite{liu2015faceattributes} on both age classification and facial recognition tasks as well as EyePACs \cite{KaggleEyePACSdataset} and CheXpert \cite{irvin2019chexpert} on disease classification. Age classification on CelebA uses accuracy and accuracy gap as its utility and fairness metrics respectively, while true positive rate and true positive rate gap are used for the remaining tasks. For facial recognition, the true positive rate is calibrated so that there is a false positive rate of $10^{-3}$ for both subpopulations. For classification tasks, we use a ResNet50 pretrained on Imagenet as our model, and adversaries are three layer multi-layer perceptrons MLP with ReLU activations. For facial recognition, we instead use the same architecture for both the model and adversaries as \cite{dhar2021pass} in using a frozen pretrained ArcFace model and appending a learnable MLP. The model takes in a image of 224 by 224 pixels for all datasets except CheXpert, which matches \cite{irvin2019chexpert} in using a resolution of 320 by 320 pixels. For training, we use alternating minimization, switching the optimization every batch for classification and every seventy batches for facial recognition. For model selection, we used the model with the best validation loss. In the grid search we perform, we train each set of $\alpha$ and $\beta$ with three different random seedings, and for visualization, we create a heatmap for each metric and dataset, grouping $\alpha$ or $\beta$ into the categories of Baseline (0, B.), Low ([0.01, 0.05], L.), Medium ([0.1-0.5], M.), and High ([1.0-10.0], H.) and taking the median in each group for use in the heatmap.

\textbf{Datasets:} CelebA is a large facial imagery datasets consisting of celebrities and providing a large number of potential attributes ranging from age to gender. We evaluate fairness with respect to gender, and use a heuristic called Individual Typology Angle (ITA), thresholded to produce a binary attribute, as a surrogate for skin color for privacy. For age classification, the task is to predict age, and for facial recognition, the task is match the identity from a database of stored features, a particular notable case where attribute privacy is important. EyePACs is a retinal imagery dataset where the task is to determine if a particular fundus photo is referable for diabetic retinopathy, where fairness is evaluated with respect to ITA again, and privacy is with respect to the quality of the fundus photo as described in \cite{Fu2019eyeq}, a surrogate for locations where imaging capabilities are less developed. CheXpert is a dataset of chest X-rays with a number of different disease classifications, of which we take predicting pleural effusion as the task. and fairness is with respect to age and privacy with respect to race.

ITA for both datasets is computed as in the procedure detailed in \cite{paul2020tara}, and both datasets for classification use training splits that exacerbate fairness by balance the task label while undersampling the positive target and sensitive attributes, while the testing split are balanced across the trio of target, sensitive, and private attributes. Splits for facial recognition are partitioned by the identity without controlling for balancing, while CheXpert reuses the provided splits.

\begin{figure*}[!t]
\centering
\begin{subfigure}{.33\textwidth}
  \centering
  \includegraphics[width=\linewidth]{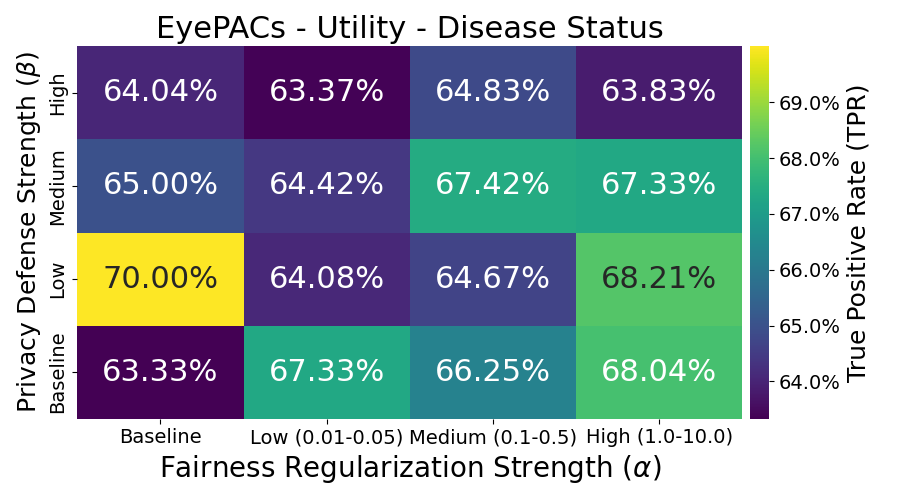}
  \label{fig:sub1}
\end{subfigure}%
\begin{subfigure}{.33\textwidth}
  \centering
  \includegraphics[width=\linewidth]{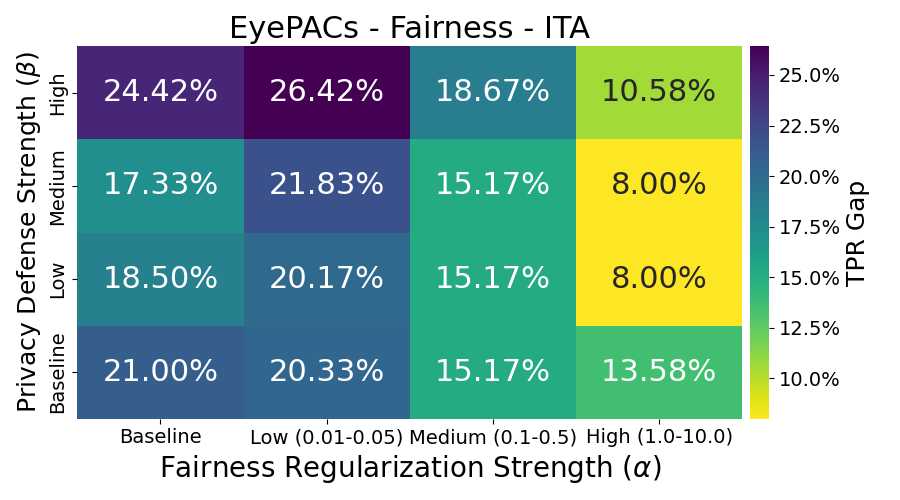}
  \label{fig:sub2}
\end{subfigure}
\begin{subfigure}{.33\textwidth}
  \centering
  \includegraphics[width=\linewidth]{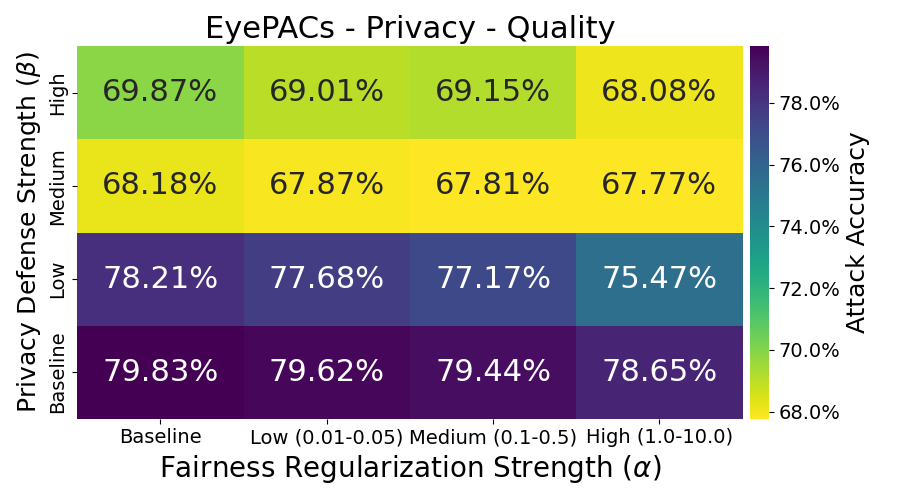}
  \label{fig:sub2}
\end{subfigure}
% \begin{subfigure}{.32\textwidth}
%   \centering
%   \includegraphics[width=\linewidth]{figures/eyepacs_test_heatmap_cai_range_0.2_0.2.png}
%   \label{fig:sub1}
% \end{subfigure}%
% \begin{subfigure}{.32\textwidth}
%   \centering
%   \includegraphics[width=\linewidth]{figures/eyepacs_test_heatmap_cai_range_0.6_0.2.png}
%   \label{fig:sub2}
% \end{subfigure}
% \begin{subfigure}{.32\textwidth}
%   \centering
%   \includegraphics[width=\linewidth]{figures/eyepacs_test_heatmap_cai_range_0.2_0.6.png}
%   \label{fig:sub2}
% \end{subfigure}
\vspace{-0.5cm}
\caption{Heatmaps on EyePACs over the different grouped regularization strengths for $\alpha$ and $\beta$.}
\label{fig:eyepacs}
\end{figure*}

\begin{figure*}[!t]
\centering
\begin{subfigure}{.33\textwidth}
  \centering
  \includegraphics[width=\linewidth]{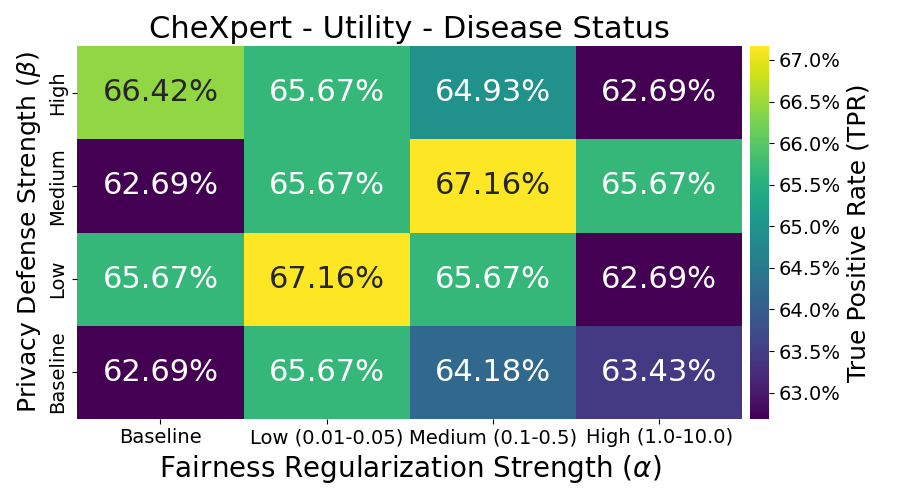}
  \label{fig:sub1}
\end{subfigure}%
\begin{subfigure}{.33\textwidth}
  \centering
  \includegraphics[width=\linewidth]{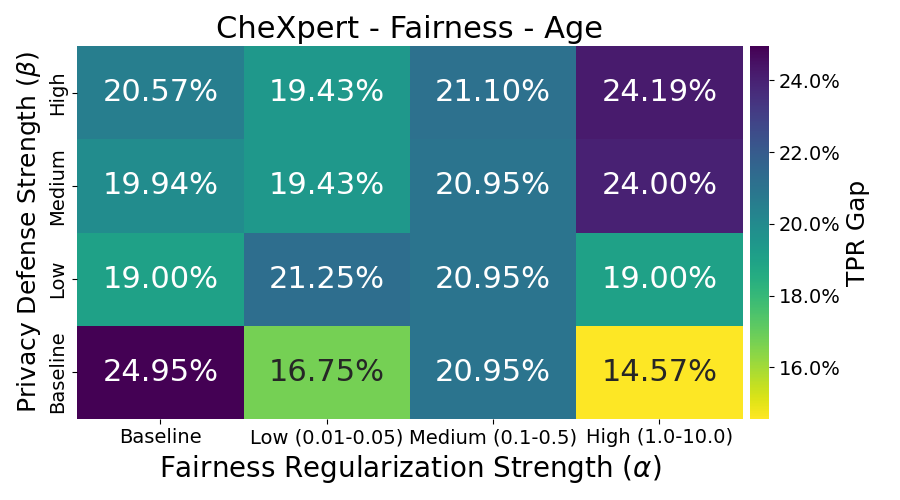}
  \label{fig:sub2}
\end{subfigure}
\begin{subfigure}{.33\textwidth}
  \centering
  \includegraphics[width=\linewidth]{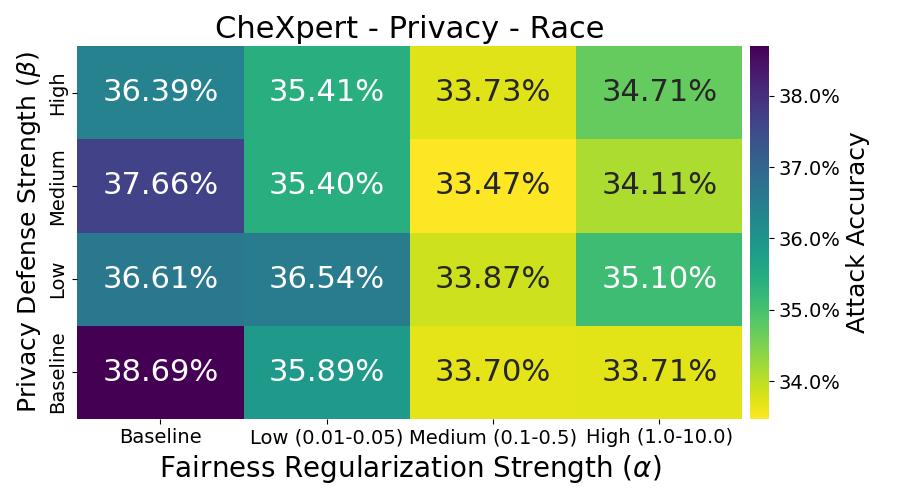}
  \label{fig:sub2}
\end{subfigure}
% \begin{subfigure}{.32\textwidth}
%   \centering
%   \includegraphics[width=\linewidth]{figures/chexpert_test_heatmap_cai_range_0.2_0.2.png}
%   \label{fig:sub1}
% \end{subfigure}%
% \begin{subfigure}{.32\textwidth}
%   \centering
%   \includegraphics[width=\linewidth]{figures/chexpert_test_heatmap_cai_range_0.6_0.2.png}
%   \label{fig:sub2}
% \end{subfigure}
% \begin{subfigure}{.32\textwidth}
%   \centering
%   \includegraphics[width=\linewidth]{figures/chexpert_test_heatmap_cai_range_0.2_0.6.png}
%   \label{fig:sub2}
% \end{subfigure}
\vspace{-0.5cm}
\caption{Heatmaps on CheXpert over the different grouped regularization strengths for $\alpha$ and $\beta$.}
\label{fig:chexpert}
\end{figure*}

\textbf{Discussion:} Tables \ref{tab:single} and \ref{tab:tradeoffs} and Figures \ref{fig:celeba}, \ref{fig:celebafr}, \ref{fig:eyepacs}, and \ref{fig:chexpert} detail our results. Starting with Table \ref{tab:single}, we see that we are able to successfully improve both fairness and privacy taken over our sweep. Improvements in utility are more muted, with medical datasets seeing larger improvements compared with the tasks on CelebA. For Table \ref{tab:tradeoffs}, the correlations taken between pairs of metrics are typically strongest between utility (U.) and fairness (U.), which is not inherently surprising given that they both deal with performance. More interesting is that only facial recognition has a strong negative correlation, where improved fairness means decreased overall performance, while classification tasks are either moderately positive or near independent. Part of this may be due to test balancing, but even for the unbalanced CheXpert test split, none of the models had worse utility than the baseline, as seen in the heatmap. Correlation between utility and privacy (P.) was more muted, with EyePACs having the highest magnitude. One caveat here, underscoring the importance of looking at all three metrics concurrently, is that for EyePACs the utility never decreased when intervening on privacy, instead increasing more with a higher $\alpha$ compared with a higher $\beta$. For the correlations between fairness and privacy, the intuition regarding the sign of the correlation matches more closely with how the trends behave, with datasets having negative correlations focusing the most on privacy typically has the worst fairness and vice versa. Finally, for the CSR metrics, we see that typically the same model is usually sufficient across the chosen preferences, with the facial recognition having the most variation, though better individual metrics can be obtained.

Regarding the figures, looking at the utility heatmap, we see that the facial recognition is the standout, with utility slightly decreasing with higher $\beta$, but much more significantly with higher $\alpha$ as noted by the correlation above. The rest of the datasets typically see little change (CelebA classification) or improvements in utility. Part of this may be due to the slightly different training procedure, though both gender and skin color are commonly considered significantly identifying features and in turn may have a more fundamental relationship to the task at hand compared to the classification based tasks. For the fairness heatmap, as one would expect we see improvements in fairness most strongly with higher levels of $\alpha$. Interesting items of note are how, without intervening on fairness directly, fairness is worsened at the highest levels of $\beta$ compared to the baseline for CelebA and EyePACs while CelebA FR and CheXpert have slightly improved fairness solely intervening on privacy, though EyePACs and CelebA see improvements at more moderate levels of $\beta$. Additionally, though intervening on both fairness and privacy for CheXpert is worse than only intervening on fairness, we see the opposite for the other datasets with the best fairness model using both interventions, suggesting that intervening on multiple attributes can be beneficial for fairness. Finally for the privacy heatmaps, we see the strongest improvements by increasing $\beta$, though, as with other cases, CheXpert is an outlier in that intervening on fairness typically produces stronger effects than intervening on privacy. We also see that the best privacy is usually acquired on a combination of both fairness and privacy, except for CelebA where the influence of the fairness regularization worsens privacy, forming a gradient from the bottom right most cell to the top left most cell.

\section{Conclusion}
This study investigates for the first time the interaction and tradeoffs between attribute privacy, fairness, and utility in computer vision and show that these interactions are more subtle and complex than straight inverse tradeoffs. The tradeoffs depend heavily on the data, with fairness and privacy at times being at odds with each other or mutually cooperative. Also, our experiments includes examples where models with the best fairness usually are some combination of addressing both fairness and privacy, suggesting to practitioners that intervening on multiple attributes can be beneficial. Potential next steps are to more tightly address these separate concerns, as well as more explicitly include given preferences in the training process.
\bibliographystyle{apalike}
\bibliography{egbib}

\end{document}